# Twin Transformer using Gated Dynamic Learnable Attention mechanism for Fault Detection and Diagnosis in the Tennessee Eastman Process

Mohammad Ali Labbaf-Khaniki*, Mohammad Manthouri**, Hanieh Ajami***

*Faculty of Electrical Engineering, K.N. Toosi University of Technology, Tehran, Iran

***Faculty of Electrical and Electronic Engineering Department, Shahed university

***Southern Illinois University Edwardsville

*mohamad95labafkh@gmail.com

**mmanthouri@shahed.ac.ir

***hajami@siue.edu

**Abstract:** Fault detection and diagnosis (FDD) is a crucial task for ensuring the safety and efficiency of industrial processes. We propose a novel FDD methodology for the Tennessee Eastman Process (TEP), a widely used benchmark for chemical process control. The model employs two separate Transformer branches, enabling independent processing of input data and potential extraction of diverse information. A novel attention mechanism, Gated Dynamic Learnable Attention (GDLAttention), is introduced which integrates a gating mechanism and dynamic learning capabilities. The gating mechanism modulates the attention weights, allowing the model to focus on the most relevant parts of the input. The dynamic learning approach adapts the attention strategy during training, potentially leading to improved performance. The attention mechanism uses a bilinear similarity function, providing greater flexibility in capturing complex relationships between query and key vectors. In order to assess the effectiveness of our approach, we tested it against 21 and 18 distinct fault scenarios in TEP, and compared its performance with several established FDD techniques. The outcomes indicate that the method outperforms others in terms of accuracy, false alarm rate, and misclassification rate. This underscores the robustness and efficacy of the approach for FDD in intricate industrial processes.

## 1) Introduction

In the modern manufacturing environment, where production is a ceaseless process, unexpected complications in control or manufacturing systems can lead to a complete shutdown of production lines. This is especially detrimental in industries that consume a lot of energy, as any downtime can result in substantial financial losses (Anitha *et al.*, 2023). Fault detection, particularly in industrial system monitoring, can utilize anomaly detection, a machine learning technique that identifies unusual data patterns, to prevent misjudgment of unknown anomalies and aid in identifying specific types of system faults (Ajami, Nigjeh and Umbaugh, 2023), and (Nigjeh, Ajami and Umbaugh, 2023). A prime example of these principles in action is the Tennessee Eastman Process (TEP), a sophisticated chemical manufacturing system (Yin *et al.*, 2012). By applying cutting-edge techniques for diagnosing and predicting faults, researchers are working to bolster the system's resilience. This ensures smooth, efficient operation and reduces the impact of unforeseen issues on production. This comprehensive approach highlights the dedication to improving fault detection methods in response to the rapidly changing manufacturing landscape (Zhang, 2022).

Transformers, a neural network architecture, have garnered considerable interest in both Natural Language Processing (NLP) and time series analysis. Their capacity to manage long-range dependencies and parallel processing has propelled their popularity in these domains (Rahali and Akhloufi, 2023). Utilizing self-attention or scaled dot-product attention, Transformers can assess the significance of inputs within a sequence, enabling the capture of intricate data patterns. In NLP, Transformers have been pivotal in achieving cutting-edge performance across tasks like translation, summarization, and sentiment analysis (Patwardhan, Marrone and Sansone, 2023). In time series prediction and classification, Recurrent Neural Network (RNN) networks have been widely used in time series detection and prediction tasks. Transformers exhibit promise, leveraging their capability to grasp temporal relationships. Applications range from weather and stock market forecasting to fault and anomaly detection. The Transformer architecture offers a significant advantage in FDD tasks by allowing the model to selectively focus on relevant sensor readings or historical data, effectively capturing complex dependencies and anomalies that may indicate faults. Additionally, the parallel computation capabilities of the Transformer enable efficient processing of large amounts of sensor data, making it well-suited for FDD applications.

The attention mechanism plays a pivotal role in the Transformer architecture, significantly enhancing its performance in tasks such as NLP and time series prediction (Vaswani *et al.*, 2017). The attention mechanism allows the model to focus on different parts of the input sequence when producing an output, effectively capturing the dependencies between words or events that are far apart (Khaniki and Manthouri, 2024). In time series prediction, the attention mechanism allows the Transformer to weigh the importance of past events when predicting future ones. Moreover, the attention mechanism in Transformers is computationally efficient as it allows for parallel computation across the sequence, unlike RNNs which require sequential computation (Samii *et al.*, 2023). In the context of FDD, the attention mechanism is particularly useful for identifying the most relevant sensor readings or historical data that contribute to the detection and diagnosis of faults. The attention mechanism in the Transformer model enhances its accuracy and robustness in FDD tasks, even with noisy or incomplete data, making it particularly valuable in industrial settings where timely and accurate fault detection is crucial for maintenance, productivity, and safety.

This research introduces a cutting-edge fault detection technique for the TEP, leveraging a dual Transformer branch that employs a unique combined attention mechanism known as Gated

Dynamic Learnable Attention (GDLAttention). The key advancements of this method are outlined below:

- **Dual Transformer Branches (Twin Transformer):** The model utilizes two distinct Transformer branches, allowing it to independently process the input data and potentially capture diverse information compared to single-branch architectures. In the simulations, this approach resulted in a better performance index such as accuracy in FDD of TEP performance compared to a baseline model without dual branches.
- **Gated Dynamic Learnable Attention (GDLAttention):** We introduce a novel attention mechanism, GDLAttention, which combines the benefits of dynamic learnable multi-head attention and learnable gate vector. By making the number of attention heads a learnable parameter, GDLAttention enables the model to adaptively adjust the number of heads during training based on the complexity of the task. The learnable gate vector, implemented through a sigmoid function, modulates the attention weights for each head, allowing the model to selectively focus on the most relevant information. This innovative approach enables the model to better capture the underlying relationships in the data and improve performance, as demonstrated in our simulations.
- **Cosine Similarity:** The attention mechanism utilizes cosine similarity, providing greater flexibility in capturing intricate relationships between query and key vectors. It disregards magnitude differences and adeptly captures nuanced feature space relationships, making it efficient for sparse data. As a result, the approach significantly improved FDD performance on the benchmark dataset compared to models using simpler similarity measures.

To assess the effectiveness of our methodology, we applied it to 21 and 18 different fault scenarios within the TEP and compared its performance with other exiting methods. We employed standard metrics, including accuracy, F1-score, precision, false alarm rate, and misclassification rate and the proposed methodology achieved significant improvements over both existing methods.

2) **FDD in Industrial Processes: Literature Review and The TEP Dataset**

FDD is vital in industrial processes to prevent damage, reduce downtime, and improve reliability. The TEP dataset is a widely-used benchmark for evaluating FDD methods, simulating a chemical processing. This section reviews existing FDD literature and introduces the TEP dataset, setting the stage for the proposed methodology.

**2.1) Literature Survey of FDD**

Machine Learning (ML) is a subset of Artificial Intelligence (AI) that enables computers to learn from data without explicit programming. Deep Learning (DL), a branch of ML, uses neural networks to mimic the human brain's learning process (Tizpaz-Niari *et al.*, 2023). Both ML and DL are essential for handling complex data and are increasingly necessary for fault diagnosis processes. Furthermore, DL has proven to be superior to traditional ML methods in many applications, achieving state-of-the-art performance in tasks such as image and speech recognition, and natural language processing. Advancements in technology drive the growing necessity to integrate cutting-edge methods ML into fault diagnosis processes. (Faizan-E-Mustafa *et al.*, 2023) explored four distinct strategies to identify and isolate faults within the TEP system using kernel Fisher discriminant analysis. (Kamal, Yu and Yu, 2014) presented a fault detection technique tailored for chemical processes that relies on independent radial basis function. (Shomal Zadeh, Khorshidi and Kooban, 2023) uses an AI methodology for fault and defect detection in concrete. The papers proposed AI-based fault detection systems by emphasizing the importance of feature extraction and understanding the relationship between physiological responses (Salehi *et al.*, 2024) and (Sanaei *et al.*, 2024).

The transformer neural network has shown great promise in fault detection applications, leveraging its self-attention mechanism to effectively identify anomalies and diagnose faults in complex systems. (Lim *et al.*, 2021) presents another adaptation of Transformer, called the temporal fusion Transformer, that combines high-dimensional and diverse inputs from multiple sources to produce accurate and interpretable forecasts for various time horizons. (Li *et al.*, 2019) proposes a novel variant of Transformer, called the time-series Transformer, that improves the performance and efficiency of Transformer on time series forecasting tasks. (Wu, Triebe and Sutherland, 2023) proposes a novel transformer-based approach for FDD in manufacturing, specifically applied to a rotary system. (Zhang *et al.*, 2022) proposes a generalized transformer-based approach FDD in the TEP, showcasing its ability to effectively detect and diagnose faults in complex industrial processes. (Du, Côté and Liu, 2023) proposes a novel method based on the self-attention mechanism containing of two diagonally-masked self-attention blocks that learn missing values from a weighted combination of temporal and feature dependencies. (Khaniki *et al.*, 2024) presents an innovative approach to FDD and cause identification by integrating multiple attention mechanisms, resulting in improved accuracy and effectiveness in TEP faults and their underlying causes. The reviewed articles suggest that transformer-based approaches have garnered significant attention from researchers in the field of fault detection and diagnosis

## 2.2) Tennessee Eastman Process

The TEP is a sophisticated and non-linear continuous process that stands as a standard in the area of fault detection, particularly in the chemical processing sector. The TEP model encompasses five main unit operations: a reactor, separator, stripper, compressor, and mixer, all of which generate both high-purity and low-purity products. The intricate and non-linear nature of the TEP, coupled with its many interdependent components, makes fault detection a considerable challenge. Faults in the TEP carry not only financial risks but also potential safety threats, underlining the essential need for prompt FDD.

The TEP dataset is a commonly used benchmark consisting of 22 datasets, 21 of which contain faults (Fault 1–21), and one (Fault 0) is without faults. Each dataset comprises 52 features or observation variables with a sampling rate of 3 minutes for most of them. The training portion of the dataset contains observations of the healthy state, while the testing portion contains faults that appear 8 hours after the training part. The dataset used in this study is derived from the TEP and is available on the Dataverse repository, which was originally used for anomaly detection research in human-automated systems. The block diagram of TEP is shown in Figure 1.

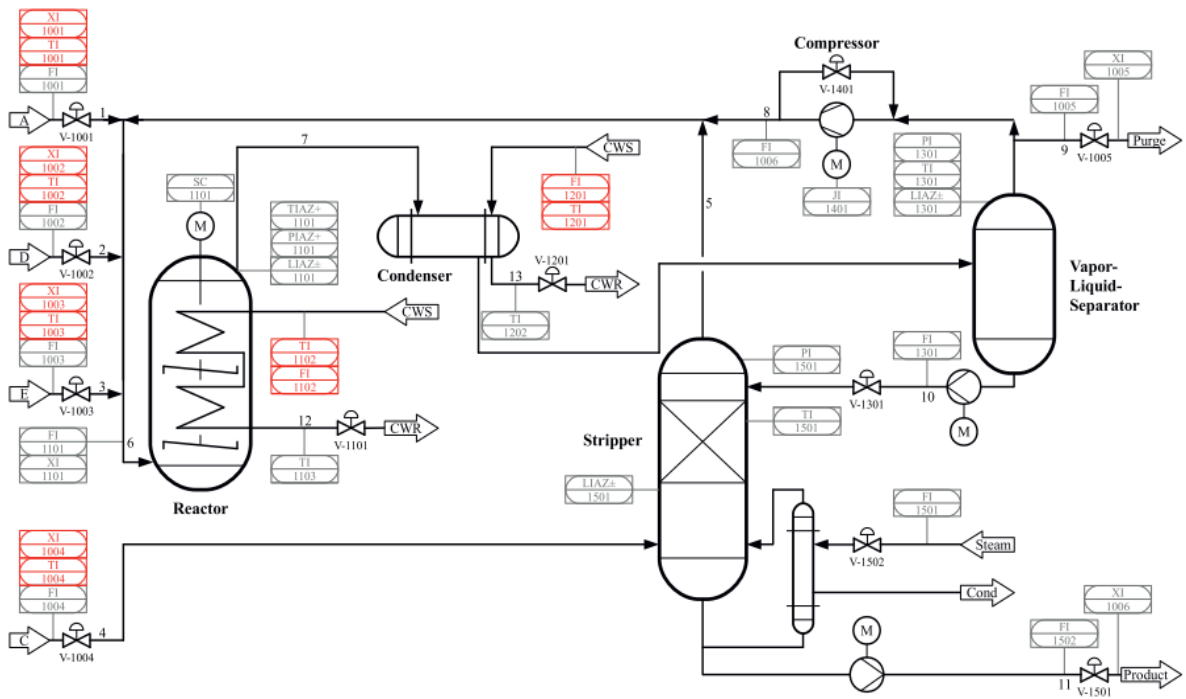

Fig. 1. The block diagram of the Tennessee Eastman Process (Reinartz, Kulahci and Ravn, 2021)

Additional details regarding the categories of each fault can be discovered in (Reinartz, Kulahci and Ravn, 2021).

## 3) Methodology

This section delves into a range of attention score methods and mechanisms, culminating in the introduction of a novel approach, GDLAttention. By integrating multiple techniques, GDLAttention enhances the ability to selectively focus on pertinent information. Its application in a twin Transformer model enables effective handling of sequential data.

### 3.1) Multi-head Attention Mechanism

Multi-head attention, a cornerstone of the Transformer architecture, enables simultaneous focus on various segments of the input sequence. The multi-head attention mechanism is a crucial component in many state-of-the-art models, particularly in handling sequential data. It captures intricate patterns and dependencies by employing multiple attention heads in parallel, allowing the model to discern temporal dependencies and relevant features in sequential data. In the context of attention mechanisms, there are three main components:

- **Queries (Q)**: These represent the information we seek from the input sequence.
- **Keys (K)**: Keys provide context about other elements in the sequence.
- **Values (V)**: Values contain the actual content associated with each element.

The process of attention involves comparing queries with keys to decide the amount of information to extract from each value. The equations for the attention scores using dot product, bilinear, and cosine methods are provided respectively.

$$A_{Dot-product} = Softmax\left(\frac{Q \cdot K^T}{\sqrt{d_k}}\right). \tag{1}$$

$$A_{Bilinear} = Softmax\left(\frac{Q \cdot W \cdot K^T}{\sqrt{d_k}}\right). \tag{2}$$

$$A_{Cosine} = Softmax\left(\frac{\frac{Q \cdot K^T}{\|Q\|\|K\|}}{\sqrt{d_k}}\right). \tag{3}$$

Here, $d_k$ denotes the dimensionality of the keys. The square root term $\sqrt{d_k}$ acts as a scaling factor. In the case of self-attention, where the queries, keys, and values all originate from the same input sequence, this mechanism enables the network to learn the optimal way to distribute information among different tokens in the sequence. This paper employs cosine similarity attention, and equation (4) illustrates the computation of this particular attention mechanism.

$$Attention(Q, K, V) = Softmax\left(\frac{\frac{Q \cdot K^T}{\|Q\|\|K\|}}{\sqrt{d_k}}\right) \times V. \tag{4}$$

In practice, the model employs several attention layers concurrently in its operation. The queries, keys, and values undergo linear projection $h$ times, each time with a different learned linear projection. This multi-layer, parallel approach allows the model to capture a richer set of features and relationships in the data. Equations of (5-6) shows the multi-head attention mechanism.

$$Multi-Head(Q, K, V) = \text{Concat}(head_1, \ldots, head_h)W_o, \tag{5}$$

$$head_i = Attention(QW_{Qi}, KW_{Ki}, VW_{Vi}) = Softmax\left(\frac{\frac{(QW_{Qi}) \cdot (KW_{Ki})^T}{\|QW_{Qi}\|\|KW_{Ki}\|}}{\sqrt{d_k}}\right) \times VW_{Vi}, \tag{6}$$

where, $W_{Qi}$, $W_{Ki}$, $W_{Vi}$ and $W_o$ are learnable parameter matrices, $h$ is the head index. These mechanisms allow the model to focus on different positions of the input sequence, and compute a weighted sum of the values, which is returned as output. The block diagram of the multi-head attention mechanism is shown in Fig. 2.

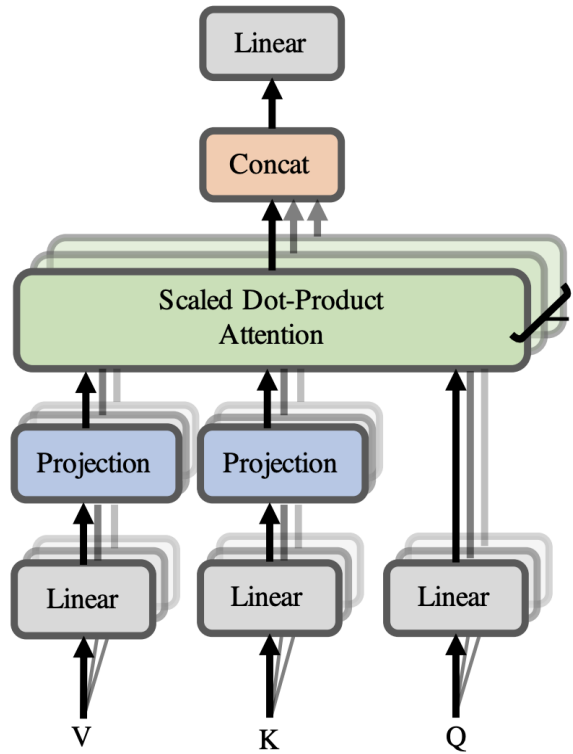

Fig. 2. The diagram of the multi-head attention mechanism

**3.2) Gated Dynamic Learnable Attention (GDLAttention)**

The GDLAttention mechanism is a pioneering innovation in attention mechanisms, introducing a novel combination of gating and dynamic learning capabilities. This innovative mechanism can be decomposed into two key components: the dynamic learnable multi-head attention mechanism and the learnable gate vector.

- Dynamic Learnable Multi-Head Attention Mechanism

In contrast to the standard multi-head attention mechanism, where the number of attention heads ($h$) is a fixed hyperparameter, our proposed dynamic learnable multi-head attention mechanism introduces a novel approach. By making the number of attention heads a learnable parameter, our model can adaptively adjust the number of heads during training based on the complexity of the task. This innovative approach enables the model to better capture the underlying relationships in the data and improve performance.

- Learnable Gate Vector

The dynamic learnable multi-head attention mechanism is achieved through the introduction of a learned gate vector $g$. Each element of the gate vector, denoted as $g_i$, modulates the attention weights for each head, allowing the model to selectively focus on the most relevant information. The output of each head is computed as:

$$head_i = g_i \cdot Attention(QW_{Qi}, KW_{Ki}, VW_{Vi}) \tag{7}$$

where $g_i$ is the $i$-th element of the learned gate vector $g$. By learning the gate vector during training, the model can dynamically adjust the number of attention heads and their corresponding weights, leading to improved performance and adaptability.

### 3.3) The Proposed Twin Transformer-GDLAttention

The core concept of the twin Transformer is the use of two parallel Transformer branches, each processing a portion of the input data independently. By processing different aspects of the data independently, the model can potentially capture a wider range of information compared to a single branch that tries to handle everything at once. Each branch can learn specialized representations tailored to the specific type of information it processes. This can lead to a more comprehensive and informative representation of the overall data. The architecture, illustrated in Fig. 3, visually represents this method's comprehensive strategy, showcasing the sequential flow of operations and the integration of twin Transformer-GDLAttention and FC layers.

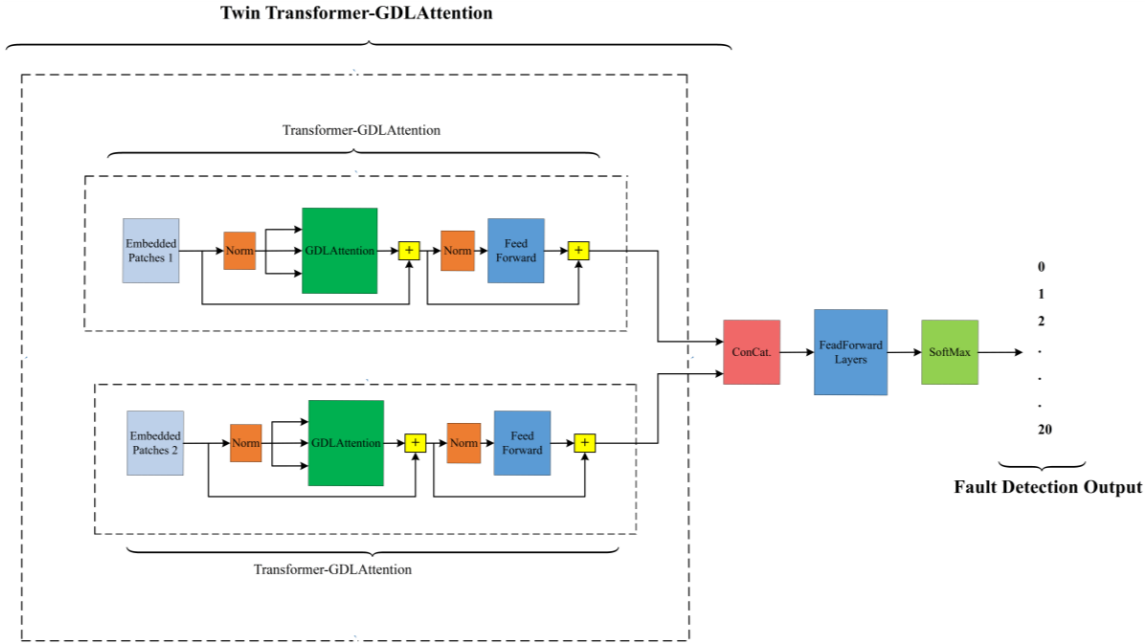

Fig. 3. The block diagram of the Twin Transformer-GDLAttention with Feedforward layer.

The TEP is a complex industrial process that poses challenges for FFD due to its non-linear dynamics and high dimensionality. Traditional LSTM-based approaches struggle to detect faults in the TEP due to their limitations in handling long-term dependencies and complex patterns. In contrast, our proposed GDLAttention approach, based on Transformer architectures, is better suited to capture long-range dependencies and nuanced patterns, enabling more accurate and efficient FDD in the TEP, particularly for detecting complex incipient faults.

The model employs Transformer blocks with a head size of 13, derived from 52/4, and is designed to work with 52 features from the TEP dataset. It utilizes 4 attention heads to capture complex dependencies within the data during initial evaluation. The feed-forward dimension is also set to 13, consistent with the attention head size. Comprising 3 sequential Transformer blocks, the model leverages multiple layers of attention and feed-forward processing to effectively enhance feature extraction and representation. Additionally, the architecture includes a feed-forward layer with a dense layer of 256 units, complemented by dropout regularization rates of 0.1 for both the Transformer and feed-forward components to mitigate overfitting. The output layer is configured for a classification task with 21 classes, making this model well-suited for multi-class sequence classification challenges. The total parameter count for this model is 85,383.

**4) Simulations**

This section details the training and validation phases of the proposed Twin Transformer-GDLAttention method for FDD in the TEP dataset. The evaluation assesses the model's performance using a range of metrics, including accuracy, precision, recall, F1-score, False Alarm Rate, and Misclassification Rate, providing a comprehensive understanding of its performance across various fault categories.

**4.1) FDD Effectiveness of the Proposed Method in terms of Accuracy**

This subsection presents the results and analysis of the proposed fault detection technique for the TEP based on the Twin Transformer-GDLAttention. The technique is compared with other methods in terms of fault classification accuracy on two different datasets: one with all faults and one without incipient faults. Fig. 4 shows the accuracy of the different FDD methods for all faults.

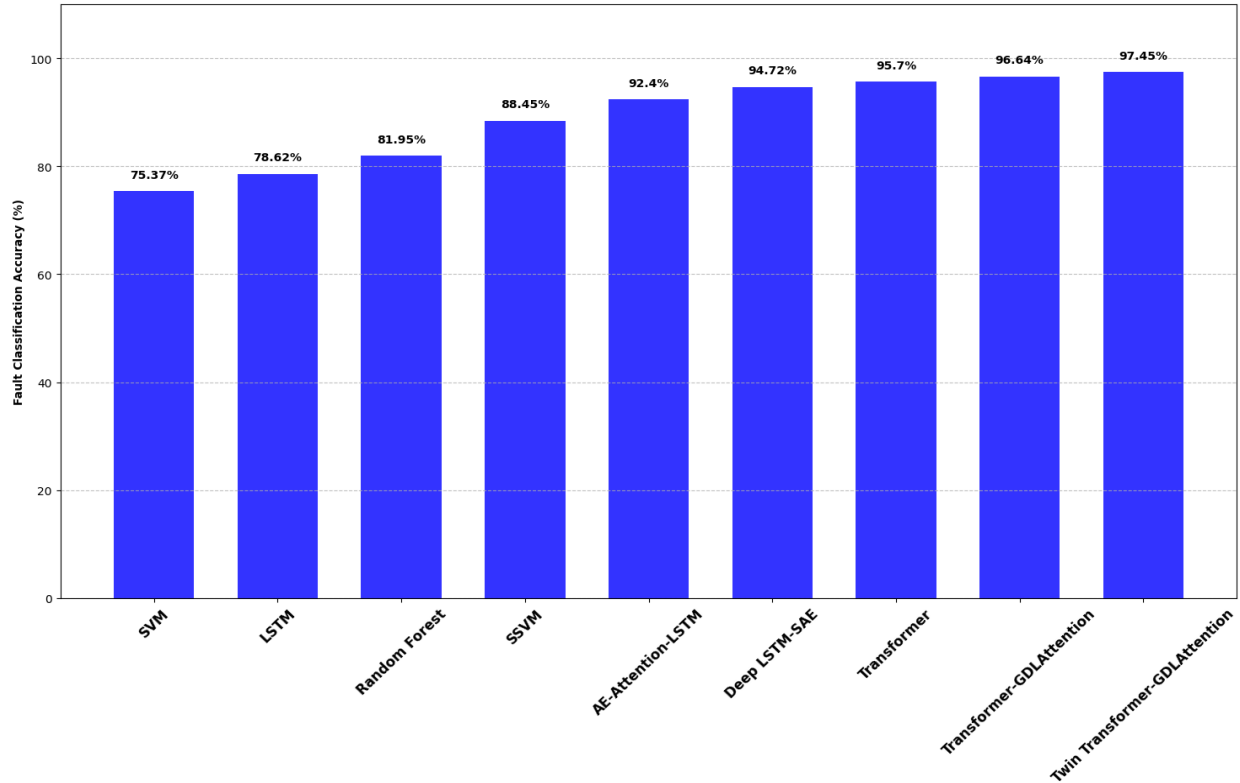

Fig. 4. Accuracy of different FDD methods for all faults

According to Fig. 4, the GDLAttention method surpasses other methods in fault classification accuracy. The proposed model's dual Transformer branches, also known as twin Transformers, process input data independently, allowing for the capture of diverse information and potentially leading to a more comprehensive understanding of the data. This, in turn, enhances performance. The GDLAttention mechanism modulates attention weights, enabling the model to selectively focus on the most relevant parts of the input, effectively filtering out less significant information. Unlike the standard multi-head attention mechanism, where the number of attention heads is a fixed hyperparameter, the gated dynamic learnable multi-head attention mechanism allows the number of attention heads to be learned during training. This adaptability enables the model to adjust the number of heads based on the task's complexity, potentially leading to improved performance. Furthermore, the attention mechanism's use of cosine similarity provides greater flexibility in capturing intricate relationships between query and key vectors, as it measures the angle between vectors regardless of their magnitude.

The proposed model's outstanding performance can be attributed to its distinctive characteristics, which distinguish it from the simple Transformer and twin Transformer. Additionally, a comparison with state-of-the-art methods, including Autoencoder-Attention-LSTM (Li, 2021) and Deep LSTM-SAE (Agarwal et al., 2022), revealed that the proposed method exhibits superior performance in FDD, indicating its enhanced effectiveness in this domain.

Faults 3, 9, and 15 are considered the most challenging to detect and diagnose in the FDD system, as they are incipient faults that exhibit subtle and complex patterns. These faults are particularly difficult to identify because their patterns are very similar to each other and to the normal operating condition of the system. This similarity makes it challenging to distinguish between the different faults and the normal condition, leading to potential misdiagnosis or delayed diagnosis. The low signal-to-noise ratio (SNR) of these faults further complicates the diagnosis process. In the case of incipient faults, the signal is often weak and can be easily masked by noise, making it difficult to detect and diagnose the fault. Moreover, the dynamic responses induced by these faults can be similar, making it challenging to determine the root cause of the fault. This similarity in dynamic responses can lead to incorrect diagnosis or delayed diagnosis, which can have serious consequences in terms of system performance, safety, and maintenance costs. Fig. 5 illustrates the accuracy of the different FDD methods for all faults except incipient faults.

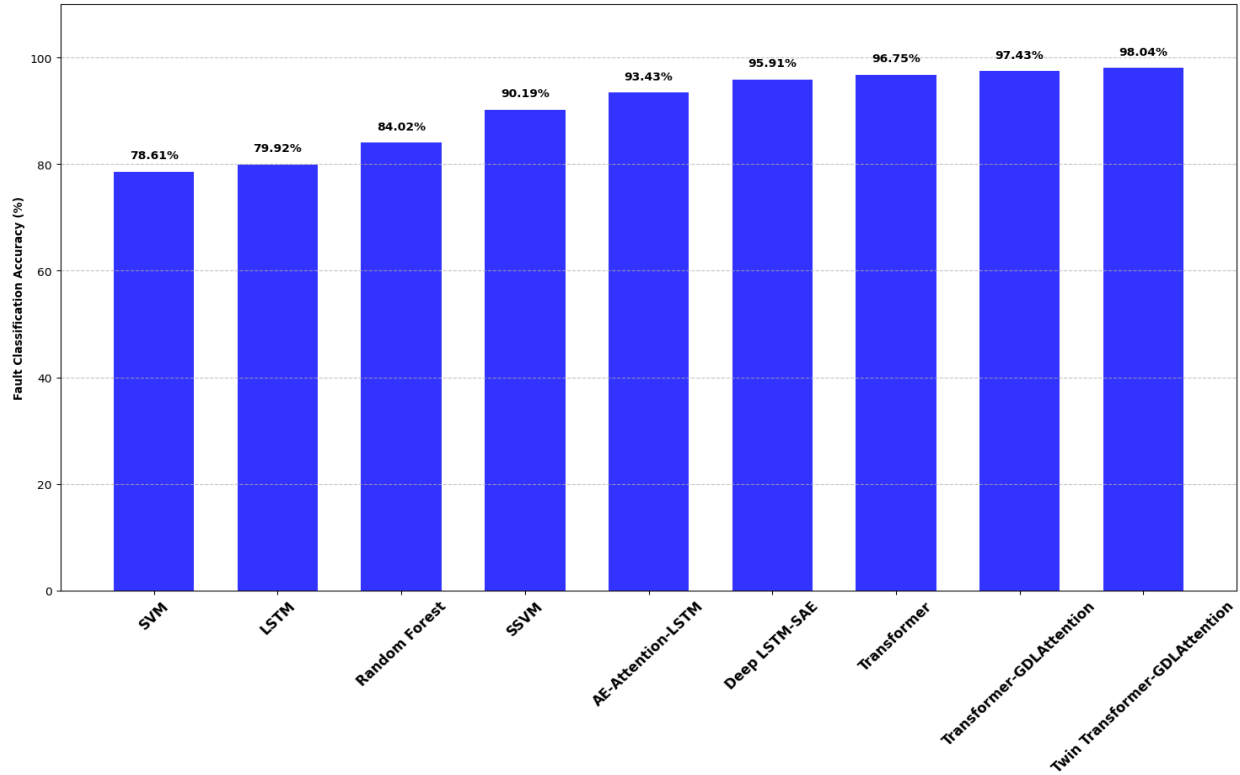

Fig. 5. Accuracy of different FDD methods for all faults except incipient faults

As evident from Fig. 5, the proposed method demonstrates superior performance compared to the simple Transformer, twin Transformer, and other state-of-the-art methods, largely due to its ability to effectively address the challenges posed by faults 3, 9, and 15. The proposed method's unique features and capabilities enable it to accurately detect and diagnose these complex faults, outperforming other methods that struggle with the subtle and complex patterns, low signal-to-noise ratio, and similar dynamic responses associated with these faults.

**4.2) FDD Performance Evaluation Using Various Metrics**

In this subsection, we embark on a comprehensive exploration of the performance indices that serve as the litmus test for the effectiveness of our proposed fault detection method. The worst-case scenarios for the FDD system are faults 3, 9, and 15, which are known to be particularly challenging to detect and diagnose (incipient faults). To evaluate the robustness of the Twin Transformer-GDLAttention system, we considered two scenarios: with incipient faults and without. To provide a comprehensive evaluation of the method's performance, we employ a range of key metrics, including Precision, Recall, F1-score, False Alarm Rate (FAR), and Missed Alarm Rate (MAR). These metrics offer a detailed insight into the method's ability to accurately identify and classify faults.

- Precision:

Precision is the ratio of correctly predicted positive observations to the total predicted positives. It provides insight into the accuracy of positive predictions.

$$Precision = \frac{TP}{TP + FN}. \qquad (8)$$

- Recall

Recall, or sensitivity/true positive rate, measures the proportion of correctly predicted positive observations to total actual positives. It evaluates the model's ability to capture all relevant positive instances.

$$Recall = \frac{TP}{TP + FN}. \qquad (9)$$

- F1-score

F1-score is the harmonic mean of precision and recall. It provides a balanced assessment of a model's precision and recall, especially useful when there is an uneven class distribution.

$$F1 - score = 2\frac{Precision \times Recall}{Precision + Recall}. \qquad (10)$$

- False Alarm Rate (FAR)

FAR, also known as false positive rate, is the ratio of false positive predictions to the total actual negatives. It measures the model's tendency to incorrectly predict the positive class.

$$FAR = \frac{FP}{TN + FP}. \tag{11}$$

- Misclassification Rate (MAR)

MAR is the ratio of incorrect predictions (both false positives and false negatives) to the total number of predictions. It provides a general overview of the model's overall accuracy.

$$MAR = \frac{FP + FN}{Total}. \tag{12}$$

Table 1 and 2 demonstrate the mentioned performance indices for two scenarios.

**Table 1.** The performance indices from the testing dataset using the Twin Transformer-GDLAttention

| Faults | Precision (%) | Recall (%) | F1-score (%) | FAR (%) | MAR (%) |
|---|---|---|---|---|---|
| Normal | 70 | 75 | 72 | 1.6 | 7.2 |
| 1 | 100 | 100 | 100 | 0 | 0.2 |
| 2 | 100 | 99 | 100 | 0 | 0.5 |
| 3 | 91 | 92 | 92 | 0.4 | 7.2 |
| 4 | 100 | 100 | 100 | 0 | 0.2 |
| 5 | 99 | 100 | 100 | 0 | 0.3 |
| 6 | 100 | 100 | 100 | 0.5 | 0.3 |
| 7 | 99 | 100 | 100 | 0.1 | 0.1 |
| 8 | 100 | 97 | 98 | 0 | 0 |
| 9 | 78 | 78 | 78 | 1.5 | 8.2 |
| 10 | 98 | 96 | 97 | 0.6 | 0.1 |
| 11 | 99 | 99 | 99 | 0.2 | 0.8 |
| 12 | 96 | 99 | 98 | 0 | 0.1 |
| 13 | 96 | 97 | 95 | 0.1 | 0.1 |
| 14 | 100 | 100 | 100 | 0.2 | 0.1 |
| 15 | 90 | 89 | 90 | 0 | 0 |
| 16 | 100 | 93 | 98 | 0.14 | 0.1 |
| 17 | 91 | 93 | 92 | 0 | 0 |
| 18 | 94 | 92 | 92 | 0.87 | 0.2 |
| 19 | 100 | 92 | 99 | 0 | 0.7 |
| **20** | 95 | 91 | 93 | 0.14 | 0.9 |
| **Average** | 95 | 94 | 94 | 0.3 | 1.3 |

**Table 2.** The performance indices from the testing dataset using the Twin Transformer-GDLAttention without the class 3, 9,15 (without the incipient faults)

| Faults | Precision (%) | Recall (%) | F1-score (%) | FAR (%) | MAR (%) |
|---|---|---|---|---|---|
| Normal | 80 | 92 | 86 | 1.3 | 7.2 |
| 1 | 100 | 100 | 100 | 0.1 | 0 |
| 2 | 100 | 99 | 99 | 0.1 | 0.1 |
| 4 | 100 | 100 | 100 | 0 | 0 |
| 5 | 100 | 100 | 100 | 0 | 0 |
| 6 | 100 | 100 | 100 | 0 | 0 |
| 7 | 100 | 100 | 100 | 0 | 0.1 |
| 8 | 100 | 97 | 97 | 0.2 | 0 |
| 10 | 97 | 95 | 97 | 0.4 | 0.1 |
| 11 | 99 | 99 | 99 | 0.2 | 0.1 |
| 12 | 96 | 99 | 98 | 0.3 | 0.1 |
| 13 | 97 | 98 | 93 | 0.1 | 0.1 |
| 14 | 100 | 100 | 100 | 0 | 0 |
| 16 | 96 | 97 | 96 | 0.4 | 0.3 |
| 17 | 100 | 94 | 96 | 0.2 | 0.1 |
| 18 | 99 | 92 | 94 | 0.5 | 0.31 |
| 19 | 99 | 99 | 98 | 0.2 | 0.1 |
| 20 | 97 | 91 | 94 | 0.41 | 0.8 |
| Average | 97 | 97 | 97 | 0.24 | 0.5 |

The results presented in Tables 1-2 confirm the effectiveness of the Twin Transformer-GDLAttention model across both scenarios, showcasing its robust performance. A summary of the key results from Tables 1-2 is presented below.

- **Precision**: With a precision of 95%, the model exhibits high accuracy in predicting positive instances, which further improves to 97% when incipient faults are excluded. This indicates that nearly all predicted positive instances are true positives.

- **Recall**: The model achieves a recall of 94%, indicating that it correctly identifies the vast majority (94%) of actual positive instances. This increases to 97% without incipient faults, showcasing the model's ability to capture most relevant positive cases.

- **F1-score**: The F1-score of 94% provides a balanced measure of the model's precision and recall, considering both false positives and false negatives. A higher F1-score indicates a more robust balance between precision and recall. This score increases to 97% when incipient faults are excluded.

- **False Alarm Rate (FAR)**: The FAR of 0.3% suggests a low rate of false positive predictions relative to the total actual negatives, indicating the model's low tendency to incorrectly predict the positive class. This decreases to 0.24% without incipient faults.

- **Misclassification Rate (MAR)**: The MAR of 1.3% indicates a low overall rate of incorrect predictions, providing a general overview of the model's accuracy. This decreases to 0.5% when incipient faults are excluded, demonstrating the model's improved performance.

### 4.3) F1-Score-Based Comparative Analysis of FDD Methods

This subsection provides a comprehensive comparison of various FDD methods, including the proposed Twin Transformer-GDLAttention method, based on their F1-scores across different fault scenarios in the TEP. Table 3 presents a comparative study, focusing on the F1-score, among various FDD methods such as Non-linear SVM (Onel, Kieslich and Pistikopoulos, 2019), CNN (Chadha and Schwung, 2017), Autoencoder-Attention-LSTM (Li, 2021), Deep LSTM-SAE (Agarwal *et al.*, 2022), simple Transformer (Wei *et al.*, 2022), and the proposed Twin Transformer-GDLAttention method.

**Table 3.** Comparison between the proposed Twin Transformer-GDLAttention methods with other FDD methods using F1-score

| Fault | PCA | Nonlinear-SVM(Onel) | LSTM | SVM | Autoencoder | CNN(Chadha and Schwung, 2017) | Autoencoder-Attention-LSTM (Li, 2021) | Deep LSTM-SAE (Agarwal et al., 2022) | Transformer(Wei et al., 2022) | Twin Transformer-GDLAttention |
|---|---|---|---|---|---|---|---|---|---|---|
| 1 | 100 | 99 | 68 | 87 | 98 | 91 | 100 | 100 | 100 | 72 |
| 2 | 79 | 98 | 74 | 88 | 85 | 88 | 89 | 100 | 98 | 100 |
| 3 | 34 | 72 | 45 | 79 | 91 | 51 | 94 | 82 | 99 | 100 |
| 4 | 99 | 100 | 75 | 90 | 89 | 100 | 99 | 100 | 100 | 92 |
| 5 | 56 | 100 | 89 | 90 | 93 | 90 | 100 | 100 | 92 | 100 |
| 6 | 99 | 100 | 1 | 95 | 85 | 91 | 100 | 100 | 99 | 100 |
| 7 | 100 | 100 | 89 | 92 | 80 | 92 | 99 | 100 | 100 | 100 |
| 8 | 97 | 95 | 71 | 63 | 73 | 83 | 99 | 100 | 96 | 100 |
| 9 | 78 | 85 | 67 | 76 | 76 | 50 | 81 | 99 | 70 | 98 |
| 10 | 66 | 96 | 77 | 89 | 79 | 70 | 99 | 42 | 98 | 78 |
| 11 | 71 | 100 | 83 | 90 | 93 | 60 | 88 | 100 | 98 | 97 |
| 12 | 99 | 91 | 56 | 75 | 91 | 86 | 99 | 100 | 97 | 99 |
| 13 | 87 | 100 | 89 | 82 | 78 | 47 | 89 | 100 | 96 | 98 |
| 14 | 98 | 65 | 99 | 88 | 1 | 89 | 99 | 100 | 99 | 95 |
| 15 | 26 | 96 | 0 | 21 | 33 | 44 | 22 | 100 | 34 | 91 |
| 16 | 24 | 93 | 0 | 14 | 89 | 67 | 31 | 100 | 53 | 90 |
| 17 | 99 | 90 | 20 | 79 | 89 | 78 | 97 | 100 | 95 | 98 |
| 18 | 78 | 88 | 88 | 66 | 95 | 83 | 89 | 100 | 94 | 92 |
| 19 | 88 | 85 | 91 | 86 | 64 | 71 | 97 | 40.4 | 98 | 100 |
| 20 | 82 | 99 | 64 | 78 | 83 | 73 | 85 | 100 | 94 | 99 |
| Ave. | 78 | 19 | 46 | 62 | 76 | 70 | 88 | 93 | 90 | 94 |
| Var. | 567 | 123 | 950 | 442 | 483 | 280 | 429 | 300 | 297 | 12 |

According to Table 3, the performance of each model is evaluated across 20 different fault scenarios in terms of F1-score, with the average and variance of their performance also provided. Comparing these results with state-of-the-art methods, it is evident that the Twin Transformer-GDLAttention model consistently outperforms the other models across most fault scenarios. It achieves an average F1-score of 94%, which is higher than the next best model, Deep LSTM-SAE (Agarwal et al., 2022), which has an average F1-score of 93%. Furthermore, the Twin Transformer-GDLAttention model exhibits the lowest variance in performance, indicating its robustness and reliability across different fault scenarios. This low variance, combined with the high average F1-score, underscores the effectiveness of the Twin Transformer-GDLAttention model. The Twin Transformer-GDLAttention model not only performs exceptionally well on average but also delivers consistent results across different scenarios. With the lowest variance in F1-scores among all the models tested, it exhibits superior stability and reliability across various classes. This balance between high performance and consistent results highlights the model's significant contribution to the field of FDD, demonstrating its overall effectiveness and reliability compared to other networks.

By including a diverse range of conventional and DL methods, our comparative study provides a comprehensive evaluation of the strengths and weaknesses of each approach. This allows us to demonstrate the robustness and effectiveness of our proposed method, Twin Transformer-GDLAttention, in FDD of TEP. Therefore, it can be concluded that the proposed comparative study provides a valuable contribution to the field, highlighting the importance of considering both conventional ML and DL methods in FDD applications. By doing so, we hope to inspire further research and development in this area, and to provide a foundation for future studies that can build upon our findings. Faults 3, 9, and 15 are the most challenging to detect and diagnose. However, the proposed method outperforms other methods in terms of F1-score for all three worst-case scenarios, demonstrating its superiority in detecting and diagnosing faults.

## 5) Conclusion

In this paper, we proposed a novel fault detection technique for the TEP based on a twin Transformer network with a GDLAttention named as Twin Transformer-GDLAttention. Our technique aimed to address the challenges of fault detection in complex and nonlinear processes, such as the TEP. The Twin Transformer architecture allows for independent processing of input data, capturing diverse information and potentially leading to a more comprehensive understanding of the data. This approach has shown to improve the FDD performance of TEP compared to single-branch architectures. The GDLAttention mechanism introduces a gating mechanism and dynamic learning capabilities. The gating mechanism, implemented through a sigmoid function, modulates the attention weights, enabling the model to focus on the most relevant parts of the input. The dynamic learning approach allows the model to adapt its attention strategy during training, potentially leading to improved performance indices. Furthermore, the attention mechanism employs cosine similarity, providing greater flexibility in capturing intricate relationships between query and key vectors. This approach has proven efficient for sparse data and has significantly improved FDD performance on the benchmark dataset compared to models using simpler similarity measures. The effectiveness of the proposed methodology was assessed across multiple fault scenarios within the TEP, and it achieved significant improvements over existing methods. Also, the proposed method outperforms other methods in detecting and diagnosing faults 3, 9, and 15, which are the most challenging to detect and diagnose, demonstrating its superiority in terms of F1-score.

Future work will focus on expanding the Twin Transformer-GDLAttention technique to other industrial processes, developing hybrid models with advanced ML methods, improving model explainability and interpretability, and exploring alternative attention mechanisms like Performer or Hierarchical Attention to enhance fault detection capabilities across diverse scenarios.

## 6) Reference

Agarwal, P. *et al.* (2022) 'Hierarchical Deep LSTM for Fault Detection and Diagnosis for a Chemical Process', *Processes*, 10(12). Available at: https://doi.org/10.3390/pr10122557.

Ajami, H., Nigjeh, M.K. and Umbaugh, S.E. (2023) 'Unsupervised white matter lesion

identification in multiple sclerosis ( MS ) using MRI segmentation and pattern classification : a novel approach with CVIPtools', 12674, pp. 1–6. Available at: https://doi.org/10.1117/12.2688268.

Anitha, C. *et al.* (2023) 'Fault Diagnosis of Tenessee Eastman Process with Detection Quality Using IMVOA with Hybrid DL Technique in IIOT', *SN Computer Science*, 4(5), p. 458. Available at: https://doi.org/10.1007/s42979-023-01851-9.

Chadha, G.S. and Schwung, A. (2017) 'Comparison of deep neural network architectures for fault detection in Tennessee Eastman process', in *2017 22nd IEEE International Conference on Emerging Technologies and Factory Automation (ETFA)*. IEEE, pp. 1–8.

Du, W., Côté, D. and Liu, Y. (2023) 'Saits: Self-attention-based imputation for time series', *Expert Systems with Applications*, 219, p. 119619.

Faizan-E-Mustafa *et al.* (2023) 'Advanced Statistical and Meta-heuristic based Optimization Fault Diagnosis Techniques in Complex Industrial Processes: A Comparative Analysis', *IEEE Access*, 11(September), pp. 104373–104391. Available at: https://doi.org/10.1109/ACCESS.2023.3317516.

Kamal, M.M., Yu, D.W. and Yu, D.L. (2014) 'Fault detection and isolation for PEM fuel cell stack with independent RBF model', *Engineering Applications of Artificial Intelligence*, 28, pp. 52–63. Available at: https://doi.org/https://doi.org/10.1016/j.engappai.2013.10.002.

Khaniki, M.A.L. *et al.* (2024) 'Enhanced Fault Detection and Cause Identification Using Integrated Attention Mechanism', *arXiv preprint arXiv:2408.00033* [Preprint].

Khaniki, M.A.L. and Manthouri, M. (2024) 'Enhancing Price Prediction in Cryptocurrency Using Transformer Neural Network and Technical Indicators', *arXiv preprint arXiv:2403.03606* [Preprint]. Available at: https://doi.org/10.48550/arXiv.2403.03606.

Li, S. *et al.* (2019) 'Enhancing the locality and breaking the memory bottleneck of transformer on time series forecasting', *Advances in neural information processing systems*, 32.

Li, Y. (2021) 'A Fault Prediction and Cause Identification Approach in Complex Industrial Processes Based on Deep Learning', *Computational Intelligence and Neuroscience*, 2021. Available at: https://doi.org/10.1155/2021/6612342.

Lim, B. *et al.* (2021) 'Temporal fusion transformers for interpretable multi-horizon time series forecasting', *International Journal of Forecasting*, 37(4), pp. 1748–1764.

Nigjeh, M.K., Ajami, H. and Umbaugh, S.E. (2023) 'Automated classification of white matter lesions in multiple sclerosis patients ' MRI images using gray level enhancement and deep learning', 12674, pp. 1–6. Available at: https://doi.org/10.1117/12.2688269.

Onel, M., Kieslich, C.A. and Pistikopoulos, E.N. (2019) 'A nonlinear support vector machine-based feature selection approach for fault detection and diagnosis: Application to the Tennessee Eastman process', *AIChE Journal*, 65(3), pp. 992–1005.

Patwardhan, N., Marrone, S. and Sansone, C. (2023) 'Transformers in the Real World: A Survey on NLP Applications', *Information*, 14(4), p. 242.

Rahali, A. and Akhloufi, M.A. (2023) 'End-to-end transformer-based models in textual-based NLP', *AI*, 4(1), pp. 54–110.

Reinartz, C., Kulahci, M. and Ravn, O. (2021) 'An extended Tennessee Eastman simulation dataset for fault-detection and decision support systems', *Computers and Chemical Engineering*, 149, p. 107281. Available at: https://doi.org/10.1016/j.compchemeng.2021.107281.

Salehi, M. *et al.* (2024) 'Innovative Cybersickness Detection: Exploring Head Movement Patterns in Virtual Reality', *arXiv preprint arXiv:2402.02725* [Preprint]. Available at: https://doi.org/10.48550/arXiv.2402.02725.

Samii, A. *et al.* (2023) 'Comparison of DEEP-LSTM and MLP Models in Estimation of Evaporation Pan for Arid Regions.', *Journal of Soft Computing in Civil Engineering*, 7(2).

Sanaei, M. *et al.* (2024) 'The Correlations of Scene Complexity, Workload, Presence, and Cybersickness in a Task-Based VR Game', *arXiv preprint arXiv:2403.19019* [Preprint]. Available at: https://doi.org/10.48550/arXiv.2403.19019.

Shomal Zadeh, S., Khorshidi, M. and Kooban, F. (2023) 'Concrete Surface Crack Detection with Convolutional-based Deep Learning Models', *International Journal of Novel Research in Civil Structural and Earth Sciences*, 10(3), pp. 25–35. Available at: https://doi.org/10.54756/IJSAR.2023.V3.10.1.

Tizpaz-Niari, S. *et al.* (2023) 'Metamorphic testing and debugging of tax preparation software', in *2023 IEEE/ACM 45th International Conference on Software Engineering: Software Engineering in Society (ICSE-SEIS)*. IEEE, pp. 138–149.

Vaswani, A. *et al.* (2017) 'Attention is all you need', *Advances in neural information processing systems*, 30.

Wei, Z. *et al.* (2022) 'A novel deep learning model based on target transformer for fault diagnosis of chemical process', *Process safety and environmental protection*, 167, pp. 480–492.

Wu, H., Triebe, M.J. and Sutherland, J.W. (2023) 'A transformer-based approach for novel fault detection and fault classification/diagnosis in manufacturing: A rotary system application', *Journal of Manufacturing Systems*, 67, pp. 439–452.

Yin, S. *et al.* (2012) 'A comparison study of basic data-driven fault diagnosis and process monitoring methods on the benchmark Tennessee Eastman process', *Journal of Process Control*, 22(9), pp. 1567–1581. Available at: https://doi.org/https://doi.org/10.1016/j.jprocont.2012.06.009.

Zhang, B. (2022) 'Rolling Bearing Fault Detection System and Experiment Based on Deep Learning', *Computational Intelligence and Neuroscience*, 2022(Vmd). Available at: https://doi.org/10.1155/2022/8913859.

Zhang, L. *et al.* (2022) 'Generalized transformer in fault diagnosis of Tennessee Eastman process', *Neural Computing and Applications*, 34(11), pp. 8575–8585.